\documentclass[11pt]{article}

\usepackage[preprint]{acl}

\usepackage{times}
\usepackage{latexsym}

\usepackage[T1]{fontenc}

\usepackage[utf8]{inputenc}

\usepackage{microtype}

\usepackage{inconsolata}

\usepackage{graphicx}
\usepackage{microtype}

\usepackage{inconsolata}
\usepackage{amsmath,graphicx}
\usepackage{tabularx}
\usepackage{amssymb}
\usepackage{amsthm}
\usepackage{cleveref}
\usepackage{pifont}
\usepackage{cancel}
\usepackage{lipsum}
\usepackage{fvextra}
\usepackage{booktabs}
\usepackage{algorithm}
\usepackage{algorithmic}
\usepackage{multirow}
\usepackage{enumitem}
\usepackage{ragged2e}
\usepackage{subcaption}
\usepackage{graphicx}
\usepackage{booktabs}
\usepackage{multirow}
\usepackage{titling}
\usepackage{pifont}
\usepackage{float}
\usepackage{xcolor}
\usepackage{graphicx}
\usepackage{booktabs}
\usepackage{multirow}
\usepackage[most]{tcolorbox}
\usepackage[table]{xcolor} 

\title{GTS: Inference-Time Scaling of Latent Reasoning with a Learnable Gaussian Thought Sampler}

\author{
Minghan Wang\textsuperscript{1}, Ye Bai\textsuperscript{2}, Thuy-Trang Vu\textsuperscript{1},\\ 
\textbf{Ehsan Shareghi}\textsuperscript{3}, \textbf{Gholamreza Haffari}\textsuperscript{1} \\
\textsuperscript{1}Department of Data Science \& AI, Monash University \\ 
\textsuperscript{2}Faculty of Medicine Dentistry and Health Sciences, University of Melbourne \\
\textsuperscript{3}Department of Computer Science, University College London \\
\texttt{\{minghan.wang,trang.vu1,gholamreza.haffari\}@monash.edu} \\
\texttt{ye.bai2@student.unimelb.edu.au} \\
\texttt{ehsan.shareghi@ucl.ac.uk} \\
}

\begin{document}
\maketitle

\begin{abstract}

Inference-time scaling (ITS) in latent reasoning models typically relies on heuristic perturbations, such as dropout or fixed Gaussian noise, to generate diverse candidate trajectories. However, we show that stronger perturbations do not necessarily yield better sampling quality: they often induce larger distribution shifts without producing more useful reasoning paths or better final decisions. A key limitation is that these perturbations inject stochasticity without defining an explicit conditional sampling distribution, making latent exploration difficult to control or optimize. To address this, we propose the Gaussian Thought Sampler (GTS), a lightweight module that reformulates latent exploration as sampling from a learned conditional distribution over continuous reasoning states. GTS predicts context-dependent perturbation distributions and is trained with GRPO-style policy optimization while keeping the backbone frozen, turning heuristic perturbation into an explicit probabilistic sampling policy. Experiments across multiple benchmarks and two latent reasoning architectures show that GTS yields more reliable inference-time scaling than heuristic baselines, suggesting that effective latent ITS requires better-controlled and optimizable sampling rather than simply amplifying stochasticity.\footnote{Code and data will be available with publication.}
\end{abstract}

\section{Introduction}

Inference-time scaling (ITS) has emerged as a central mechanism for enhancing the reasoning performance of large language models (LLMs). By allocating additional test-time compute to generate and select among multiple reasoning trajectories, approaches such as self-consistency and best-of-$N$ sampling significantly improve accuracy without modifying model parameters~\citep{wang2023selfconsistencyimproveschainthought,wang2024mathshepherdverifyreinforcellms,cobbe2021trainingverifierssolvemath,DBLP:conf/iclr/LightmanKBEBLLS24}. In discrete, token-based LLMs, such scaling is naturally supported by explicit conditional distributions over next tokens. Sampling strategies such as temperature scaling or nucleus sampling operate directly on these distributions, implicitly trading off diversity and likelihood under a probabilistic framework.

Recent advances in continuous latent reasoning introduce a different computational regime. These models perform multi-step reasoning directly in hidden state space, refining latent thought representations without generating intermediate textual tokens \citep{hao2024traininglargelanguagemodels, shen2025codicompressingchainofthoughtcontinuous,sui2025stopoverthinkingsurveyefficient,zhu2025surveylatentreasoning}. While this paradigm improves reasoning efficiency and expressivity, it also removes the explicit token-level probability distributions that enable principled sampling in discrete models. As a result, ITS in latent reasoning models typically relies on heuristic perturbations, such as dropout or injecting fixed Gaussian noise~\citep{wang2025inferencetimescalingcontinuousspace,you2026paralleltesttimescalinglatent}. These perturbations introduce stochasticity, but they do not define an explicit sampling distribution over latent thoughts.

\begin{figure*}[t]
    \centering
    \includegraphics[width=1.0\textwidth]{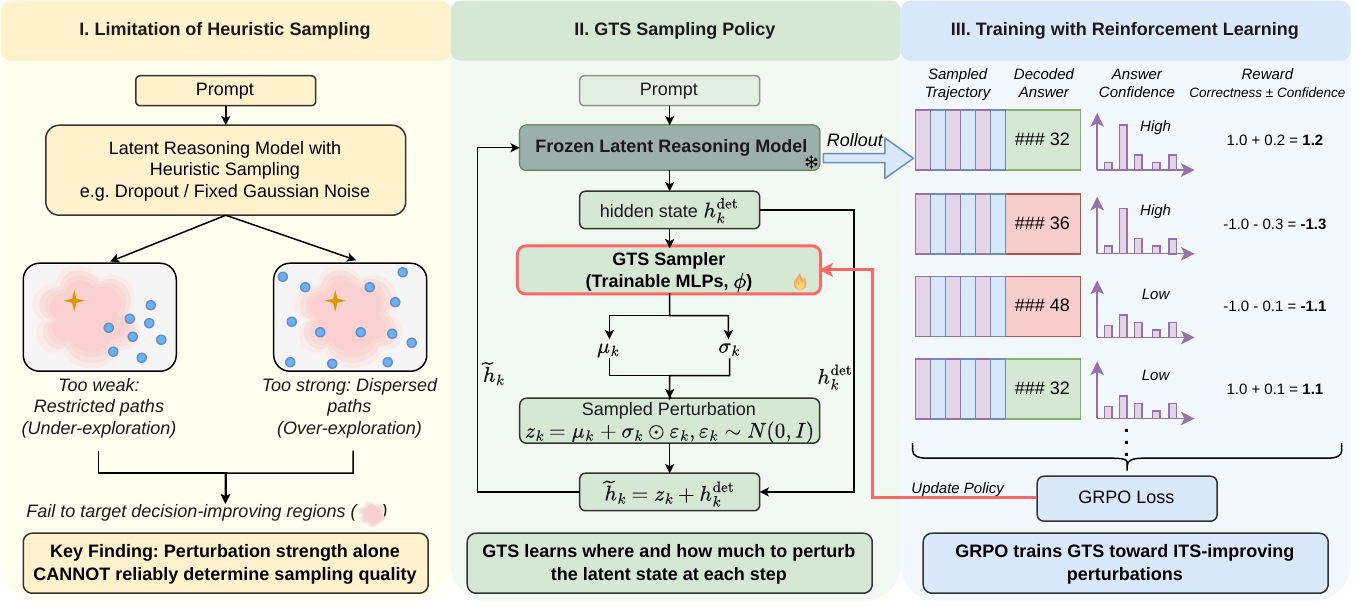}
    \caption{
    Overview of \textbf{GTS}. \textbf{Left:} Heuristic latent perturbations cannot reliably control sampling quality: weak noise under-explores, while strong noise over-disperses trajectories. \textbf{Middle:} GTS predicts a context-dependent Gaussian perturbation distribution from the deterministic latent state at each step. \textbf{Right:} Trajectories are scored by correctness and confidence, and GRPO trains GTS to favor perturbations that improve inference-time scaling.
    }
    \label{fig:GTS}
    \vspace{-1em}
\end{figure*}

We therefore revisit sampling in ITS through the lens of \emph{exploration over reasoning trajectories} at inference time\footnote{We formalize this notion of exploration in \S\ref{sec:pre_study:exploration_define}.}, which raises a fundamental question: \textbf{what constitutes effective exploration in continuous reasoning space?} Our analysis (\S\ref{sec:pre_study} and \S\ref{sec:analysis}) reveals that the key challenge in latent ITS is not simply how much perturbation to inject, but how exploration is properly controlled. Existing methods, such as dropout or fixed Gaussian noise, rely on manually chosen perturbation scales that are highly sensitive to the backbone, task, and latent geometry, making them prone to under- or over-exploration (i.e., overly concentrated or excessively diffuse search). Moreover, these methods mainly alter sampling diversity, whereas effective ITS depends not only on diversity, but also on whether sampling shifts the model's belief toward regions that better support the correct answer, 
as illustrated in~\Cref{fig:GTS}. 
Finally, heuristic perturbations do not define an explicit conditional sampling distribution, leaving latent thought sampling as a hand-crafted stochastic process rather than a principled, optimizable policy.

To address this limitation, we reformulate latent exploration as conditional sampling over continuous thought representations. Specifically, we introduce a \textbf{Gaussian Thought Sampler} (GTS), a lightweight module that goes beyond fixed noise scaling by predicting context-dependent Gaussian perturbation distributions over latent reasoning states. This turns latent exploration from a heuristic perturbation mechanism into an explicit probabilistic sampling policy, bringing latent ITS closer to the distributional perspective that underlies token-level inference-time scaling. Across multiple latent reasoning backbones and benchmarks, GTS consistently improves scaling performance over dropout-based and standard Gaussian perturbations, showing that effective latent ITS requires not just more stochasticity, but better-guided sampling. Our contributions are as follows:
\begin{itemize}[leftmargin=*,itemsep=1pt,topsep=1pt]
    \item We identify a key limitation of heuristic latent ITS: effective scaling requires controlling sampling quality, not just injecting stochasticity.
    \item We propose GTS to reformulate latent perturbation as conditional sampling, turning heuristic perturbation into an explicit and optimizable sampling policy.
    \item Across multiple backbones and benchmarks, we show that this principled sampling view consistently outperforms dropout-based and standard Gaussian perturbations.
\end{itemize}
\section{Diagnostic Analysis of Heuristic Sampling}
\label{sec:pre_study}

\subsection{Exploration as Trajectory Sampling in ITS}
\label{sec:pre_study:exploration_define}

ITS improves reasoning by allocating additional test-time compute to generate multiple candidate trajectories and then selecting or aggregating their resulting answers~\citep{wang2023selfconsistencyimproveschainthought}. This procedure is closely related to \emph{exploration} in reinforcement learning (RL): during rollout, a stochastic policy samples multiple trajectories, and performance improves when exploration increases the chance of reaching a better outcome~\citep{DBLP:books/lib/SuttonB98,stiennon2022learningsummarizehumanfeedback,deepseekai2025deepseekr1incentivizingreasoningcapability}. Motivated by this parallel, we formalize ITS as \textbf{exploration over reasoning trajectories} at inference time. Given an input $x$, let $\mathcal{T}(x)$ denote the set of valid reasoning trajectories for $x$, and let $\pi(\tau \mid x)$ be the stochastic policy used in the inference-time sampling procedure. Under an \textbf{exploration budget} of $N$, ITS draws
\begin{equation}
\tau^{(1)}, \dots, \tau^{(N)} \sim \pi(\cdot \mid x), \qquad \tau^{(i)} \in \mathcal{T}(x),
\end{equation}
and produces the final prediction by selecting or aggregating the answers decoded from these trajectories. Unlike RL, ITS does not update $\pi$ using reward feedback; exploration is used only to generate candidate trajectories, while performance gains arise from selecting among the sampled outcomes.

This perspective also clarifies the role of sampling in different reasoning regimes. In text-based reasoning models, $\pi(\tau \mid x)$ is directly induced by token-level stochastic decoding from explicit conditional distributions, i.e., by sampling discrete actions from a categorical policy over the vocabulary at each reasoning step. In latent reasoning models, in contrast, intermediate reasoning steps are typically generated by deterministic state transitions,
\begin{equation}
h_{t+1}^{\mathrm{det}} = f_{\theta}(h_t^{\mathrm{det}}, x),
\end{equation}
which do not directly define a stochastic trajectory distribution (where $f_{\theta}$ is the LLM and $h^{\mathrm{det}}$ is the deterministic hidden state). Existing methods, therefore, introduce stochasticity by perturbing latent dynamics, for example
\begin{equation}
\tilde{h}_{t+1} = f_{\theta}(\tilde{h}_t, x) + z_t,
\end{equation}
where $z_t$ is a stochastic perturbation. Such perturbations induce trajectory sampling in latent space and thereby enable exploration.

Under finite budgets, the central issue is not merely whether exploration produces different trajectories, but whether it produces \emph{useful} ones. We therefore focus on \textbf{exploration quality}: whether the induced sampling policy tends to generate reasoning trajectories that move the model's predictive belief toward the correct answer, rather than merely introducing random variation. This leads to the core question of this section: \textbf{do heuristic perturbations provide effective exploration in latent reasoning models, or do they mostly add stochasticity without improving the chance of finding a correct answer?} To answer this, we isolate and evaluate sampling quality independently of end-task accuracy.

\subsection{Experimental Setup}

\paragraph{Models}
We evaluate a text-based reasoning model, i.e., GPT-2~\citep{radford2019language} fine-tuned on GSM8K-Aug~\citep{deng2023implicitchainthoughtreasoning}, and a latent reasoning model, \textsc{COCONUT}, built on the same backbone. GPT-2 produces textual reasoning followed by the delimiter "\texttt{\#\#\#}`` before answer generation. \textsc{COCONUT} performs $K=6$ latent reasoning steps, following prior implementation details~\citep{hao2024traininglargelanguagemodels}.
\vspace{-1.5mm}
\paragraph{Sampling Protocol}
For each input $x$ in the GSM8K-test~\citep{cobbe2021trainingverifierssolvemath} dataset, we generate $N=32$ reasoning trajectories by applying sampling only to the reasoning stage. For each trajectory $\tau$, we quantify the predictive probability of the first ground-truth answer token $y_1^\star$ using teacher forcing after appending the answer prefix. Since GSM8K answers are numeric, we focus on $y_1^\star$ to reduce multi-token noise while preserving decision information. GPT-2 uses token-level sampling (temperature $1.0$) and dropout sampling ($p \in \{0.1,0.5\}$). \textsc{COCONUT} applies dropout-based perturbations in latent space.

\subsection{Measuring Sampling Quality}

Let $\tau^{\mathrm{det}}$ denote the deterministic reasoning trajectory, and let $p(y_1^\star \mid x,\tau)$ be the predictive probability of the first correct answer token.

\paragraph{Sampling Gain (SG)}
We define trajectory-level gain as the change in log-odds of the correct answer:
\begin{align}
\Delta(\tau)
&=
s(\tau)
-
s(\tau^{\mathrm{det}}),
\\
s(\tau)
&=
\log \frac{p(y_1^\star \mid x,\tau)}
{1 - p(y_1^\star \mid x,\tau)}
\end{align}
Positive $\Delta(\tau)$ indicates improved decision confidence relative to the deterministic baseline. To reflect best-of-$N$ selection, we define
\begin{equation}
\mathrm{SG}(x) = \max_{k \le N} \Delta(\tau_k),
\end{equation}
and report the dataset-level mean SG.

\paragraph{Sampling Gain Rate}
We additionally report the fraction of inputs with $\mathrm{SG}(x) > 0.5$, corresponding to a substantial increase in decision odds.

\paragraph{Distribution Shift}
We quantify how much sampling shifts the answer distribution using the Jensen--Shannon divergence~(JS) between answer-token distributions:
\begin{equation}
\mathrm{JS}(\tau)
=
\mathrm{JS}\!\left(
p(\cdot \mid x,\tau)
\;\|\;
p(\cdot \mid x,\tau^{\mathrm{det}})
\right).
\end{equation}
Mean JS reflects how strongly sampling alters the model’s answer distribution, independent of whether such changes are beneficial.

\subsection{Results}

\Cref{tab:prelim} reveals a consistent pattern across both models. For GPT-2, token sampling achieves the highest SG and SG rate while inducing only moderate JS. This suggests that probabilistic token sampling can improve the probability of the correct answer while keeping distribution shift moderate under finite budgets.
Increasing dropout strength produces a different effect. While mild dropout ($p=0.1$) can approximate token sampling, stronger dropout ($p=0.5$) substantially increases JS but sharply reduces SG. Larger perturbations therefore induce greater distribution shift without reliably improving sampling quality.
The effect is more pronounced in \textsc{COCONUT}. Mild latent dropout yields positive SG with minimal JS, whereas stronger dropout leads to negative SG despite increased divergence.

Overall, higher distribution shift does not necessarily imply better sampling quality. Across both models, stronger perturbations consistently increase divergence from the deterministic trajectory, but this does not translate into higher SG and can even reduce it. This mismatch is especially pronounced in latent reasoning, where heuristic perturbations are more likely to move sampling away from decision-improving regions. These results suggest that effective latent ITS requires more than simply increasing stochasticity, and instead calls for better-controlled sampling mechanisms.

\begin{table}[t]
\centering
\setlength{\tabcolsep}{4pt}
\renewcommand{\arraystretch}{0.92}
\begin{tabularx}{\columnwidth}{@{}>{\raggedright\arraybackslash}X c c c@{}}
\toprule
\textbf{Sampling} & \textbf{SG} $\uparrow$ & \textbf{SG $>$ 0.5} $\uparrow$ & \textbf{JS} \\
\midrule

\rowcolor{gray!12}
\multicolumn{4}{c}{\textbf{GPT-2}} \\
Token (temp=1.0)  & $\mathbf{9.94}$ & $\mathbf{0.62}$ & $0.29$ \\
Dropout ($p=0.1$) & $9.64$          & $0.58$          & $0.30$ \\
Dropout ($p=0.5$) & $3.89$          & $0.57$          & $0.67$ \\
\midrule

\rowcolor{gray!12}
\multicolumn{4}{c}{\textbf{\textsc{COCONUT}}} \\
Dropout ($p=0.1$) & $\mathbf{1.09}$ & $\mathbf{0.61}$ & $0.05$ \\
Dropout ($p=0.5$) & $-0.87$         & $0.40$          & $0.28$ \\
\bottomrule
\end{tabularx}

\caption{Preliminary sampling quality analysis on GPT-2 and \textsc{COCONUT}. We report SG, SG rate, and distribution shift (JS). Higher is better for SG and SG rate.}
\vspace{-5mm}
\label{tab:prelim}
\end{table}

\section{Gaussian Thought Sampler}
\label{sec:method}

\subsection{Overview}

We augment a frozen latent reasoning backbone with a learnable, context-conditioned sampling policy over hidden-state perturbations. Instead of injecting heuristic noise into deterministic latent reasoning, GTS models exploration as an explicit conditional density over continuous perturbations. This reformulates inference-time exploration as a probabilistic modeling problem and enables direct optimization of the sampling mechanism.

\subsection{Conditional Latent Sampling}
\label{sec:method:model}

\paragraph{Problem Setup}
Given an input question $x$, a latent reasoning model parameterized by $\theta$ performs $K$ latent reasoning steps and then produces an answer distribution:
\begin{equation}
p_\theta(y \mid x, \mathbf{h}_{1:K})
\end{equation}
where $\mathbf{h}_k \in \mathbb{R}^d$ denotes the backbone hidden state at step $k$. Our goal is to introduce a learnable conditional sampling distribution over latent perturbations while keeping $\theta$ fixed.

\vspace{-1.5mm}
\paragraph{Perturbation Variable}
We introduce a continuous perturbation variable $\mathbf{z}_k \in \mathbb{R}^d$ at each reasoning step and define:
\begin{equation}
\tilde{\mathbf{h}}_k
=
\mathbf{h}_k^{\mathrm{det}} + \mathbf{z}_k,
\label{eq:tildeh_new}
\end{equation}
where $\mathbf{h}_k^{\mathrm{det}}$ is the deterministic hidden state. The perturbed state $\tilde{\mathbf{h}}_k$ is fed back into the backbone for subsequent reasoning, yielding a sampled latent trajectory: $\tau = \{\tilde{\mathbf{h}}_1, \dots, \tilde{\mathbf{h}}_K\}$.

\paragraph{Context-Conditioned Gaussian Policy}
We parameterize a conditional Gaussian policy over perturbations:
\begin{equation}
q_\phi(\mathbf{z}_k \mid \mathbf{c}_k)
=
\mathcal{N}\!\big(
\boldsymbol{\mu}_\phi(\mathbf{c}_k),
\mathrm{diag}(\boldsymbol{\sigma}_\phi^2(\mathbf{c}_k))
\big),
\label{eq:qz_new}
\end{equation}
where $\mathbf{c}_k$ denotes the conditioning context at step $k$ (in practice, the backbone hidden state $\mathbf{h}_k^{\mathrm{det}}$). Sampling follows the reparameterization:
\begin{equation}
\mathbf{z}_k
=
\boldsymbol{\mu}_\phi(\mathbf{c}_k)
+
\boldsymbol{\sigma}_\phi(\mathbf{c}_k)
\odot
\boldsymbol{\epsilon}_k,
\quad
\boldsymbol{\epsilon}_k \sim \mathcal{N}(\mathbf{0},\mathbf{I}).
\label{eq:reparam_new}
\end{equation}

Because~\Cref{eq:tildeh_new} defines an affine transformation with unit Jacobian, the change of variables preserves density. Therefore, learning $q_\phi(\mathbf{z}_k \mid \mathbf{c}_k)$ is equivalent to learning an explicit conditional density over perturbed thought representations $\tilde{\mathbf{h}}_k$. The backbone computation remains unchanged, while GTS governs how perturbations are sampled during latent reasoning.

\subsection{Policy Learning}
\label{sec:method:train_new}

GTS defines an explicit conditional density over latent perturbation trajectories. 
A natural alternative would be likelihood-based training, e.g., ELBO-style variational objectives~\citep{kingma2022autoencodingvariationalbayes}. 
However, our goal is not to model a latent posterior, but to directly optimize inference-time exploration under task-level rewards. 
The quality of a perturbation trajectory is determined by ITS performance, which are non-differentiable with respect to the sampler parameters.
We therefore treat latent perturbations as continuous actions and optimize the policy via reinforcement learning.

\vspace{-1.5mm}
\paragraph{Trajectory Policy}

For an input $x$, a perturbation trajectory 
$\tau=\{\mathbf{z}_1,\dots,\mathbf{z}_K\}$ 
defines a factorized Gaussian policy:
\begin{equation}
\log q_\phi(\tau \mid x)
=
\sum_{k=1}^K
\log q_\phi(\mathbf{z}_k \mid \mathbf{c}_k).
\label{eq:traj_density_final}
\end{equation}
For a diagonal Gaussian, the per-step log-density admits a closed-form expression:
\begin{align}
&\log q_\phi(\mathbf{z}_k \mid \mathbf{c}_k)
= \\ \nonumber
&-\frac{1}{2}
\sum_{d=1}^{D}
\Big[
\big(\tfrac{z_{k,d}-\mu_{k,d}}{\sigma_{k,d}}\big)^2
+
2\log\sigma_{k,d}
+
\log(2\pi)
\Big].
\label{eq:logq_gauss_final}
\end{align}
This closed-form density is essential: it enables exact computation of policy likelihoods and density ratios, making policy-gradient optimization well-defined in continuous latent space.


\paragraph{Reward Design}

For each input $x$, we sample $N$ trajectories and decode answers $\{a^{(i)}\}_{i=1}^N$ during rollout. 
Let $y^\star$ denote the ground-truth answer. 
The reward for trajectory $i$ is defined as
\begin{equation}
r^{(i)}
=
r_0\big(2\,\mathbb{I}[a^{(i)}=y^\star]-1\big)
+
\alpha\, s^{(i)},
\label{eq:reward_final}
\end{equation}
where the first term provides a symmetric correctness signal and 
$s^{(i)}$ is a confidence-based shaping term derived from the normalized log-probability of the generated answer. 
The shaping term encourages high-confidence correct trajectories and discourages high-confidence incorrect ones, while remaining secondary to correctness. See Appendix~\ref{app:reward_shaping} for more details.

\paragraph{GRPO-Style Policy Optimization}

To stabilize policy updates, we maintain a reference sampler 
$q_{\phi_{\mathrm{ref}}}$ as an exponential moving average of the current policy~\citep{schulman2017proximalpolicyoptimizationalgorithms,ouyang2022traininglanguagemodelsfollow}. 
For trajectory $\tau^{(i)}$, we compute the density ratio:
\begin{equation}
\rho^{(i)}
=
\frac{
q_\phi(\tau^{(i)} \mid x)
}{
q_{\phi_{\mathrm{ref}}}(\tau^{(i)} \mid x)
}.
\label{eq:ratio_final}
\end{equation}
Following GRPO-style clipped optimization~\citep{deepseekai2025deepseekr1incentivizingreasoningcapability}, the policy gradient objective is:
\begin{align}
&\mathcal{L}_{\mathrm{PG}}
= \\ \nonumber
&-\mathbb{E}
\Big[
\min\big(
\rho^{(i)} A^{(i)},
\mathrm{clip}(\rho^{(i)},1-\epsilon_c,1+\epsilon_c)\, A^{(i)}
\big)
\Big],
\label{eq:pg_final}
\end{align}
where $A^{(i)}$ denotes the group-normalized advantage computed within each prompt, and $1\pm\epsilon_c$ defines the clipping range. We further regularize the sampler via a KL penalty between the current and reference Gaussian policies:
\begin{equation}
\mathcal{L}_{\mathrm{KL}}
=
\beta\,
\mathbb{E}
\big[
\mathrm{KL}\big(
q_\phi(\mathbf{z}_k\mid\mathbf{c}_k)
\|
q_{\phi_{\mathrm{ref}}}(\mathbf{z}_k\mid\mathbf{c}_k)
\big)
\big],
\label{eq:kl_final}
\end{equation}
which also admits a closed-form solution. The final objective is:
\begin{equation}
\mathcal{L}_{\mathrm{GTS}}
=
\mathcal{L}_{\mathrm{PG}}
+
\mathcal{L}_{\mathrm{KL}}.
\label{eq:gts_final}
\end{equation}
This formulation directly optimizes the inference-time exploration distribution in continuous latent space while leaving the base language model unchanged.

\section{Experiments}

\begin{figure*}[t]
    \centering
    \resizebox{\textwidth}{!}{
        \begin{tabular}{cc}
            \includegraphics[width=0.48\textwidth]{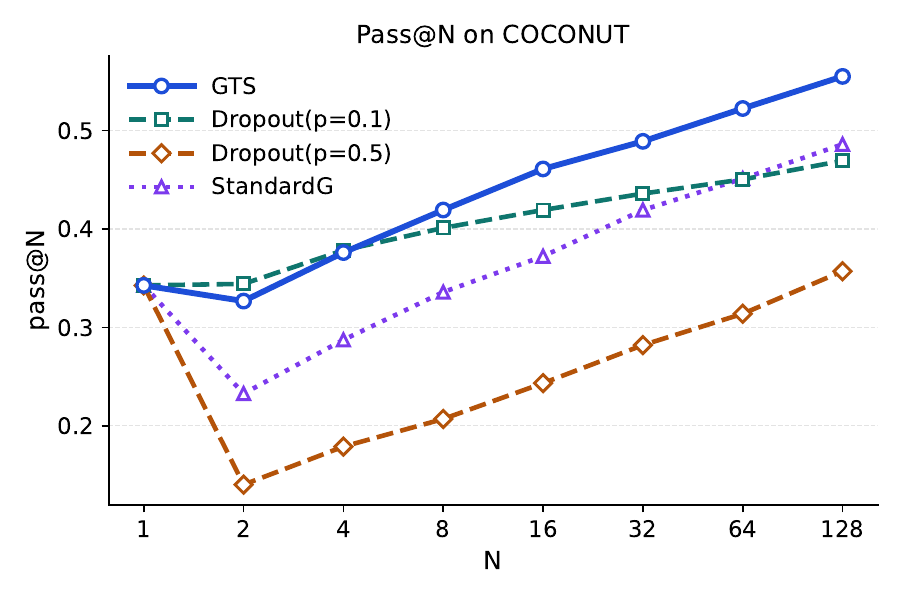} &
            \includegraphics[width=0.48\textwidth]{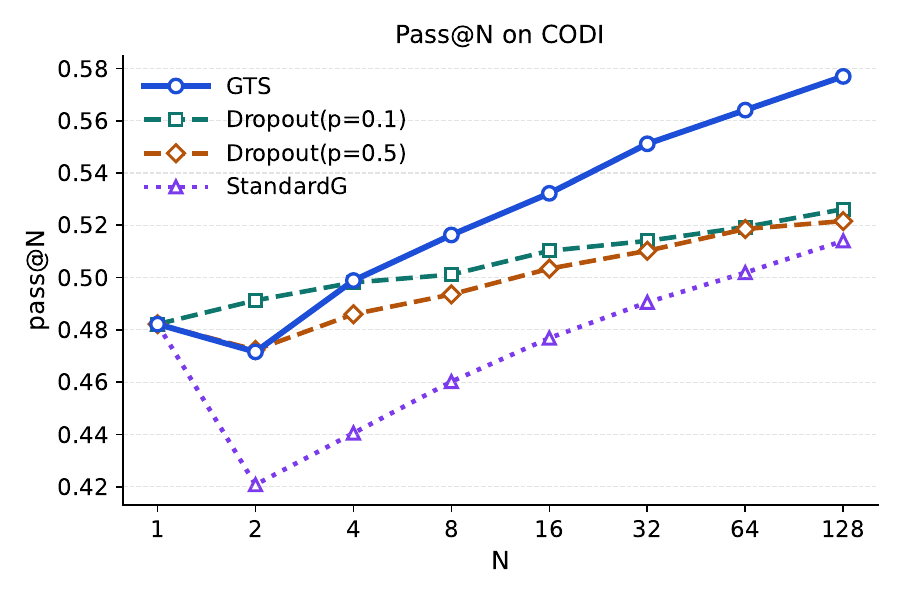}
            \vspace{-5mm}
        \end{tabular}
    }
    \vspace{-1mm}
    \caption{
    \textbf{ITS performance under different sampling strategies.}
    Pass@N on \textsc{COCONUT} (left) and \textsc{CODI} (right) as a function of sampling budget $N$.
    All methods coincide at $N=1$, corresponding to deterministic latent reasoning.
    As $N$ increases, \textbf{GTS} achieves stronger scaling behavior than dropout-based sampling and standard Gaussian noise (StandardG), indicating more effective exploration of the latent reasoning space. More results of our out-of-distribution evaluation on the other 3 benchmarks can be found in~\S\ref{app:ood_evaluation}.
    }
    \vspace{-4mm}
    \label{fig:latent_pan}
\end{figure*}

\subsection{Experimental Setup}
\label{sec:exp:setup}

\paragraph{Data}

We use the \textbf{GSM8K-aug} training corpus adopted in prior latent reasoning work~\citep{deng2023implicitchainthoughtreasoning}. 
From the full augmented set (386k), we uniformly sample 20k training instances and train all samplers for one epoch. 
Evaluation is performed on the standard GSM8K~\cite{cobbe2021trainingverifierssolvemath} test set (1319 samples). To further assess generalization beyond the training distribution, we additionally evaluate the same trained samplers on three out-of-distribution arithmetic reasoning benchmarks: \textbf{MultiArith}~\citep{roy-roth-2015-solving}, \textbf{SVAMP}~\citep{patel-etal-2021-nlp}, and \textbf{GSM8K-Hard}~\citep{gao2023palprogramaidedlanguagemodels}. No additional training or hyperparameter tuning is performed for these datasets; all samplers are applied directly at inference time under the same settings as the main experiments.

\paragraph{Models}

We evaluate GTS on two latent reasoning models: \textsc{COCONUT}~\citep{hao2024traininglargelanguagemodels} and \textsc{CODI}~\citep{shen2025codicompressingchainofthoughtcontinuous}. 
For \textsc{COCONUT}, we follow the architecture and protocol described in \citet{hao2024traininglargelanguagemodels}, using a GPT-2 backbone with $K=6$ latent reasoning steps. 
For \textsc{CODI}, we use a \textsc{LLaMA-3.2-1B}~\citep{grattafiori2024llama3herdmodels} backbone with 6 latent reasoning steps and its recurrent filtering module. In all cases, backbone parameters are frozen and only the Gaussian sampler is trained.

\paragraph{GTS Architecture}

The sampler consists of lightweight mean and log-standard-deviation heads parameterizing a diagonal Gaussian policy over latent perturbations. 
To avoid premature collapse to a deterministic policy, we enforce a minimum log-standard-deviation during training ($\log \sigma > -2.0$).
Latent perturbations are injected only during recursive latent reasoning steps (from \texttt{<start\_latent>} to \texttt{<lat\_5>}), ensuring that stochasticity only affects the reasoning dynamics rather than the final aggregation stage.
Additional implementation details are provided in Appendix~\ref{app:implementation_details}.

\paragraph{Training}

For each prompt, we sample $N=32$ perturbation trajectories during rollout.
Unless otherwise stated, training runs for 10K optimization steps with batch size set as 32 and a learning rate of $1\times10^{-4}$ with linear warmup. 
Other hyperparameters can be found in Appendix~\ref{app:implementation_details}.
\vspace{-1.5mm}
\paragraph{Baselines}
We compare GTS against two stochastic inference baselines:
\begin{itemize}[leftmargin=*,itemsep=1pt,topsep=1pt]
    \item \textbf{Dropout Sampling} 
    Dropout ($p \in \{0.1, 0.5\}$) is enabled during latent reasoning steps while remaining disabled during prompt prefilling and answer generation.
    \item \textbf{Standard Gaussian Noise}
    At each latent step, we add isotropic Gaussian noise 
    $\boldsymbol{\epsilon}\sim\mathcal{N}(\mathbf{0},\mathbf{I})$
    to the hidden state without context conditioning or learned parameters.
\end{itemize}

\vspace{-1.5mm}
\paragraph{Evaluation}
We evaluate ITS performance by varying the sampling budget 
$N\in\{1,2,4,8,16,32,64,128\}$. 
When $N=1$, all methods reduce to deterministic inference. 
For $N\ge2$, stochasticity is applied exclusively to latent reasoning, and answers are decoded greedily to isolate the effect of thought sampling. Our primary metric is $\mathrm{pass@}N$.

\subsection{Main Results}
\label{sec:exp:result}

\paragraph{Overall scaling behavior}

\Cref{fig:latent_pan} shows pass@N curves on \textsc{COCONUT} and \textsc{CODI}. As the sampling budget $N$ increases, all stochastic methods improve over deterministic inference, confirming that sampling latent thoughts can enable ITS. However, these gains are not strictly monotonic at very small budgets. In particular, GTS and the standard Gaussian noise baseline (StandardG) can slightly reduce performance at $N=2$, suggesting that a small amount of stochasticity may initially disrupt trajectory quality before the benefit of multi-sample selection becomes apparent.

\paragraph{Comparison with baselines}
Heuristic perturbation methods show substantially different behaviors across settings. Dropout with a mild rate ($p=0.1$) provides stable gains on both models and remains a relatively strong baseline, especially at small-to-moderate budgets. In contrast, stronger dropout ($p=0.5$) consistently performs worse, most clearly on \textsc{COCONUT}, where it causes a sharp small-budget degradation and remains well below the other methods throughout. StandardG is also less stable, with weaker small-budget behavior and less consistent scaling than mild dropout. Taken together, these results suggest that fixed heuristic perturbations can help, but their effectiveness is highly sensitive to perturbation strength and model dynamics.

\begin{figure*}[t]
    \centering
    \resizebox{\textwidth}{!}{
        \begin{tabular}{cc}
            \includegraphics[width=0.48\textwidth]{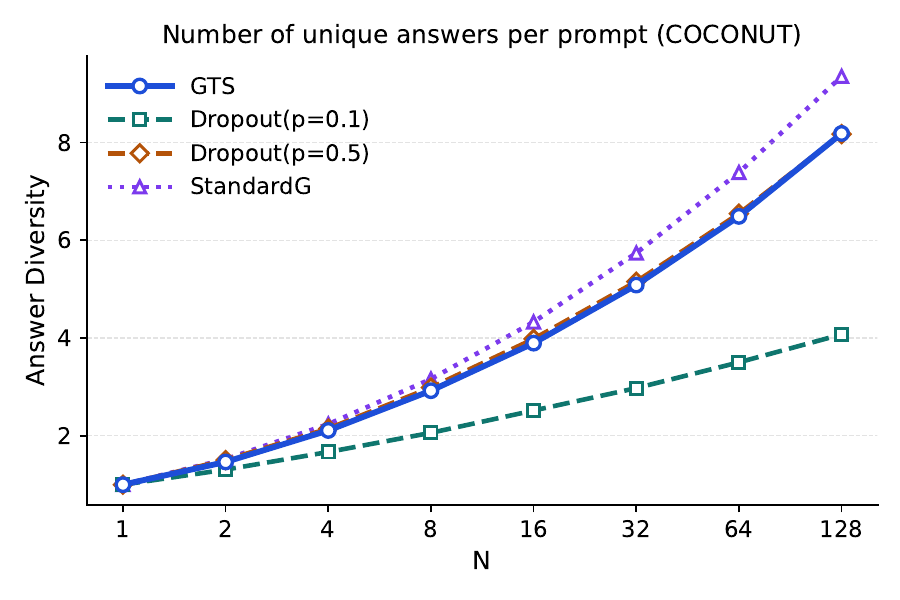} &
            \includegraphics[width=0.48\textwidth]{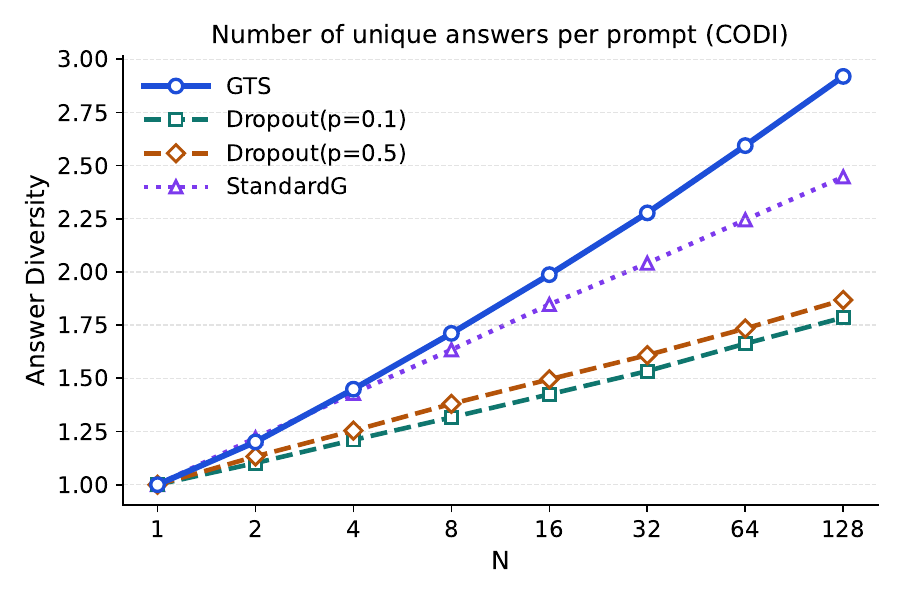}
            \vspace{-5mm}
        \end{tabular}
    }
    \caption{
    Average number of unique decoded answers per prompt.}
    \vspace{-3mm}
    \label{fig:latent_div}
\end{figure*}

\paragraph{Effectiveness of GTS}
GTS achieves the strongest scaling behavior at moderate-to-large budgets on both backbones, and is the best-performing method overall. Compared with heuristic baselines, its advantage becomes clearer as $N$ grows, indicating that the learned sampler produces more useful trajectory diversity under finite inference budgets. The relative gain is larger on \textsc{COCONUT} than on \textsc{CODI}, suggesting that the benefit of learned perturbation control may depend on the latent reasoning dynamics of the backbone; we provide further analysis in Appendix~\ref{app:further_main_results}. In addition, from Appendix~\ref{app:ood_evaluation}, out-of-distribution results on three benchmarks show the same overall pattern, with GTS remaining more consistent than heuristic perturbations. Overall, these results show that learning a context-conditioned perturbation policy yields more effective latent ITS than globally fixed stochastic perturbations.

\section{Analysis}
\label{sec:analysis}

\subsection{On Sampling Quality}
\label{sec:analysis_sample_quality}

\begin{table}[t]
\centering
\setlength{\tabcolsep}{4pt}
\renewcommand{\arraystretch}{0.92}
\begin{tabularx}{\columnwidth}{@{}>{\raggedright\arraybackslash}X c c c@{}}
\toprule
\textbf{Sampling} & \textbf{SG} $\uparrow$ & \textbf{SG $>$ 0.5} $\uparrow$ & \textbf{JS} \\
\midrule

\rowcolor{gray!12}
\multicolumn{4}{c}{\textbf{\textsc{COCONUT}}} \\
Dropout ($p=0.1$) & $1.09$          & $0.61$          & $0.05$ \\
Dropout ($p=0.5$) & $-0.87$         & $0.40$          & $0.28$ \\
StandardG         & $0.34$          & $0.51$          & $0.20$ \\
\textbf{GTS}      & $\mathbf{1.84}$ & $\mathbf{0.82}$ & $0.11$ \\
\midrule

\rowcolor{gray!12}
\multicolumn{4}{c}{\textbf{\textsc{CODI}}} \\
Dropout ($p=0.1$) & $0.53$          & $0.31$          & $0.01$ \\
Dropout ($p=0.5$) & $0.47$          & $0.39$          & $0.06$ \\
StandardG         & $0.11$          & $0.38$          & $0.15$ \\
\textbf{GTS}      & $\mathbf{1.15}$ & $\mathbf{0.70}$ & $0.10$ \\
\bottomrule
\end{tabularx}
\caption{Sampling quality analysis for latent reasoning models on GSM8K. 
We report SG, SG rate, and distribution shift (measured by JS divergence). Higher is better for SG and SG rate.}
\vspace{-3mm}
\label{tab:analysis_sampling_quality}
\end{table}

We revisit sampling quality using the diagnostic metrics introduced in~\S\ref{sec:pre_study}.
Following the same protocol and dataset, we evaluate SG, SG rate, and distribution shift on both \textsc{COCONUT} and \textsc{CODI}, together with {StandardG} as a Gaussian perturbation baseline consistent with the ITS setup in~\Cref{sec:exp:setup}.

\paragraph{Failure modes of heuristic perturbations}

\Cref{tab:analysis_sampling_quality} shows that the mismatch between distribution shift and sampling quality observed in \S\ref{sec:pre_study} persists across both latent reasoning backbones. We summarize this behavior with two recurring regimes:
\begin{itemize}[leftmargin=*,itemsep=1pt,topsep=1pt]
\item \textbf{Under-exploration} refers to perturbations that are too weak to move sampling meaningfully away from the deterministic trajectory, yielding limited distribution shift and only modest gains.
\item \textbf{Over-exploration} refers to perturbations that are too strong and disrupt decision-relevant information, producing large distribution shift but limited or even negative SG, together with reduced SG rate.
\end{itemize}
These regimes are clearly reflected in~\Cref{tab:analysis_sampling_quality}. For \textsc{COCONUT}, mild dropout remains in the under-exploration regime, while stronger dropout moves into over-exploration, sharply increasing distribution shift but driving SG negative. For \textsc{CODI}, the same pattern appears more mildly: stronger perturbations again increase distribution shift without proportional gains in sampling quality. Across both backbones, GTS avoids these extremes and achieves positive SG with relatively controlled distribution shift. Overall, effective latent ITS depends not simply on increasing stochasticity, but on keeping perturbations within a useful regime. Heuristic methods do not control this trade-off explicitly, whereas GTS learns a context-dependent perturbation policy that balances diversity and decision improvement more reliably under finite budgets.

\subsection{On Sampling Behavior}

\paragraph{Answer-level diversity}

\Cref{fig:latent_div} reports the average number of unique decoded answers per prompt in the main experiment.
Compared to distribution shift in \Cref{tab:analysis_sampling_quality}, which measures changes in the answer distribution, answer diversity captures discrete branching behavior after greedy decoding. Higher diversity does not necessarily imply stronger sampling gain or better pass@$N$ scaling.
In particular, StandardG often produces more unique decoded answers than GTS at moderate budgets, yet this additional branching does not translate into higher SG or stronger pass@N scaling. This again shows that more diverse outputs are not necessarily more decision-useful ones.

\paragraph{Step-wise signal-to-noise ratio}

\begin{figure}[t]
    \centering
    \includegraphics[width=\columnwidth]{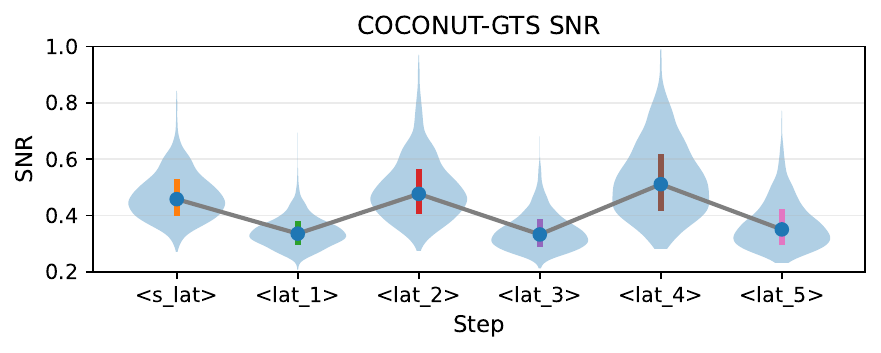}
    \includegraphics[width=\columnwidth]{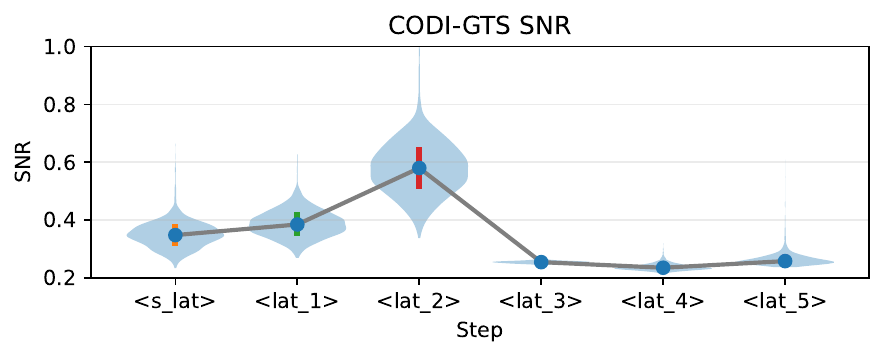}
    \caption{
    Step-wise distribution of signal-to-noise ratio (SNR) across latent reasoning steps.
    Each violin shows the distribution over prompts; markers denote medians.
    }
    \vspace{-4mm}
    \label{fig:snr}
\end{figure}

To analyze how stochastic intervention evolves across latent reasoning steps, we measure a step-wise signal-to-noise ratio (SNR).
At latent step $t$, the sampler predicts a mean vector $\mu_t \in \mathbb{R}^D$ and diagonal log standard deviation $\log \sigma_t \in \mathbb{R}^D$. We define:
\begin{equation}
    \text{SNR}_t =
\frac{
\sqrt{\frac{1}{D} \lVert \mu_t \rVert_2^2}
}{
\sqrt{\frac{1}{D} \sum_{i=1}^D \sigma_{t,i}^2}
}.
\end{equation}
SNR measures the relative strength of deterministic steering versus injected noise.
Values below 1 indicate noise-dominated steps, while larger values indicate stronger deterministic influence.
We evaluate on GSM8K-test, sampling $N=32$ trajectories per prompt and averaging SNR at the prompt level.
Dataset-level distributions are shown in~\Cref{fig:snr}.

For \textsc{COCONUT}, SNR alternates across adjacent latent steps, forming a saw-tooth pattern with moderate magnitudes.
Deterministic and stochastic components remain comparable throughout reasoning, indicating interleaved refinement and variation.
For \textsc{CODI}, deterministic strength concentrates early (peaking at \texttt{<lat\_2>}) and remains lower in later steps, suggesting a more front-loaded adjustment pattern. Overall, GTS does not impose a fixed stochastic schedule. Instead, the balance between deterministic and stochastic components adapts to the latent reasoning dynamics of the underlying backbone. See more discussions in Appendix~\ref{app:snr_discussion}.

\subsection{Ablation on Reward Shaping}


\begin{figure}
    \centering
    \includegraphics[width=\columnwidth]{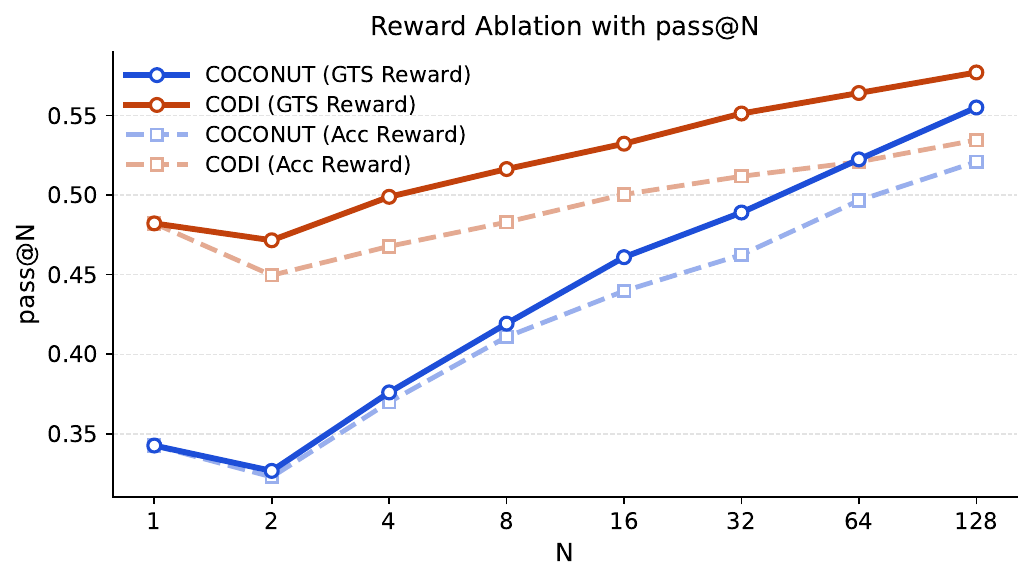}
    \vspace{-8mm}
    \caption{Ablation study on the reward shaping. Pass@N of the GTS trained with the accuracy-only reward and the dense reward introduced in~\S\ref{sec:method:train_new}.}
    \label{fig:reward_ablation_pan}
    \vspace{-4mm}
\end{figure}

We compare GTS trained with the full dense reward in~\Cref{sec:method:train_new} against an accuracy-only variant.
In the simplified setting, the shaping term is removed by setting $\alpha=0$, reducing the reward to $r^{(i)} = \{-1, 1\}$.
All other training configurations remain identical.
\Cref{fig:reward_ablation_pan} shows pass@$N$ curves on GSM8K-test for both \textsc{COCONUT} and \textsc{CODI}.
The dense reward consistently outperforms the accuracy-only variant, and the performance gap widens as $N$ increases. The shaping term effectively provides additional within-group discrimination, improving the quality of sampled candidate sets.
In contrast, the accuracy-only reward treats all correct (and incorrect) samples equally, limiting refinement during training.

\section{Related Work}

\subsection{ITS in Discrete Space}

ITS improves reasoning by allocating additional test-time compute to generate and select among multiple reasoning paths. A representative approach is self-consistency~\citep{wang2023selfconsistencyimproveschainthought}, which samples multiple Chain-of-Thought (CoT) solutions and aggregates them via majority voting. This idea has been extended through best-of-$N$ sampling and reranking, where candidate trajectories are scored by likelihood, confidence signals, or external verifiers.

Beyond unstructured sampling, structured prompting frameworks introduce explicit search over the discrete reasoning space. Least-to-Most~\citep{zhou2023leasttomostpromptingenablescomplex}, Tree-of-Thoughts~\citep{yao2023treethoughtsdeliberateproblem}, and Graph-of-Thoughts~\citep{Besta_2024} formulate reasoning as systematic exploration over branching intermediate states.

Another complementary direction develops Verifier or Process Reward Models (PRMs) to evaluate intermediate reasoning steps. Math-Shepherd~\citep{wang2024mathshepherdverifyreinforcellms} automatically generates step-level supervision from CoT outputs, while subsequent work improves robustness and generalization of process-level feedback~\citep{zhang2025lessonsdevelopingprocessreward}. OpenPRM~\citep{zhang2025openprm} further extends process supervision to open-domain settings through preference-based evaluation. Collectively, these methods rely on explicit token-level distributions and scoring signals, making exploration and selection relatively controllable in discrete space.

\subsection{Continuous Space Reasoning}

Continuous CoT reasoning performs multi-step inference directly in latent space, refining hidden representations without emitting intermediate textual tokens~\citep{sui2025stopoverthinkingsurveyefficient}. By operating on continuous manifolds, this paradigm aims to improve reasoning efficiency and representational expressivity~\citep{zhu2025surveylatentreasoning}, e.g., CoT2~\citep{gozeten2025continuouschainthoughtenables} demonstrates that LLMs can maintain multiple reasoning traces in parallel within continuous states.

Most existing work focuses on learning stable and compact latent representations during training. CODI~\citep{shen2025codicompressingchainofthoughtcontinuous} aligns student and teacher hidden states via self-distillation, while CCOT~\citep{cheng2024compressedchainthoughtefficient} introduces variable-length latent embeddings with optional decoding for interpretability. Hybrid approaches such as Token Assorted~\citep{su2025tokenassortedmixinglatent} combine discrete tokens with latent reasoning. \textsc{COCONUT}~\citep{hao2024traininglargelanguagemodels} further shows that complex reasoning can be executed primarily within hidden state space.

While these works advance latent representation learning, they largely assume static inference passes. Systematically scaling test-time computation within continuous manifolds remains relatively under-explored.

\subsection{ITS in Continuous Space}

Recent efforts begin to explore ITS directly in continuous space. One direction promotes diversity in latent trajectories. For example, SoftCoT++~\citep{xu2025softcottesttimescalingsoft} generates multiple ``soft thoughts'' from distinct initial tokens using contrastive objectives. Another direction samples and aggregates multiple trajectories. CoT2~\citep{gozeten2025continuouschainthoughtenables} represents parallel reasoning paths as superpositions of continuous tokens, while \citet{zhang2026silencejudgereinforcementlearning} employ self-verification signals based on proximity to a latent centroid. \citet{wang2025inferencetimescalingcontinuousspace,you2026paralleltesttimescalinglatent} introduce Monte Carlo Dropout to induce stochasticity and aggregate sampled trajectories with a learned reward model.

Despite these advances, most existing approaches rely on heuristic perturbations, such as dropout or fixed Gaussian noise to induce diversity. Because such stochasticity is not explicitly conditioned on semantic context, its magnitude is difficult to calibrate and may shift sampling away from decision-relevant regions, particularly under larger sampling budgets. Similar limitations have been noted in prior analysis~\citep{wang2025inferencetimescalingcontinuousspace}.

To address this gap, we propose GTS, which reformulates latent perturbation as conditional sampling from an explicit, learnable Gaussian distribution over latent representations. By modeling exploration through a parameterized density, GTS enables explicit and optimizable test-time exploration, providing a principled alternative to heuristic noise injection.

\section{Conclusion}
We study inference-time scaling in latent reasoning models through the lens of conditional sampling in continuous thought space. Our analysis shows that heuristic perturbations do not reliably produce effective exploration: larger distribution shift or higher answer diversity does not necessarily translate into better sampling quality, and fixed perturbation schemes can easily fall into under- or over-exploration. To address this limitation, we introduce GTS, a lightweight Gaussian sampler that models latent perturbation as an explicit, context-conditioned sampling policy. Across two latent reasoning architectures, GTS yields stronger and more reliable scaling under finite budgets than heuristic baselines. Overall, our results suggest that effective latent ITS requires not just more stochasticity or diversity, but better-controlled sampling that more reliably supports correct final decisions.
\section*{Limitations}
This work has several limitations. Although we evaluate GTS beyond the training distribution on multiple arithmetic reasoning benchmarks, our empirical scope remains limited to relatively short, answer-focused math tasks and does not yet cover more open-ended, long-form, or non-mathematical reasoning settings. We restrict the sampling policy to a diagonal Gaussian distribution and do not explore broader perturbation families that may offer different flexibility-stability trade-offs. Our analysis of sampling behavior remains empirical and does not provide a formal theoretical characterization of exploration in high-dimensional latent spaces. Finally, we study two representative latent reasoning architectures, and the behavior of learnable perturbation policies may differ under alternative continuous reasoning formulations. We leave broader task coverage, distributional extensions, and theoretical analysis to future work.


\bibliography{custom}

\clearpage
\appendix

\section*{Appendix}
\label{sec:appendix}

\section{Additional Experimental Results}
\begin{figure*}[t]
    \centering
    \resizebox{\textwidth}{!}{
        \begin{tabular}{cc}
            \includegraphics[width=0.48\textwidth]{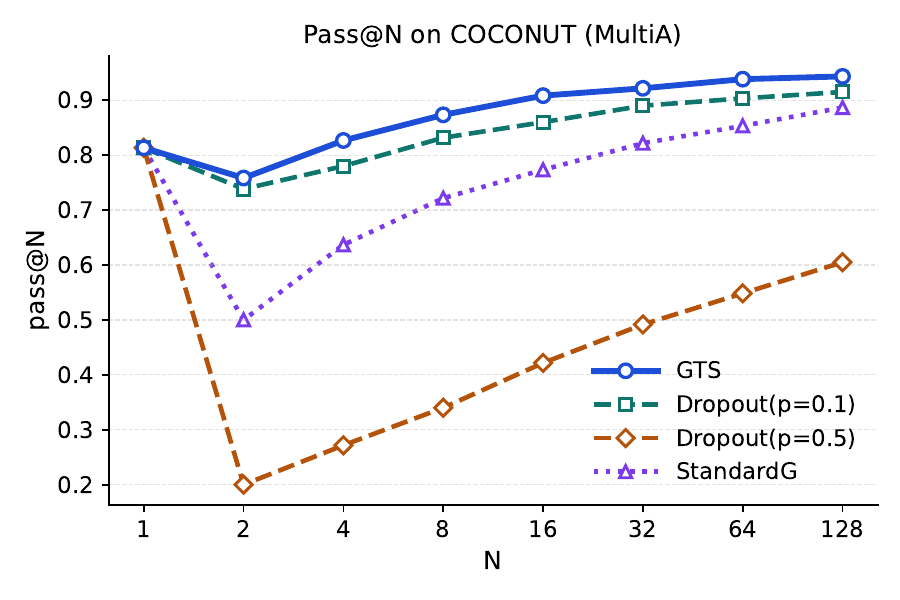} &
            \includegraphics[width=0.48\textwidth]{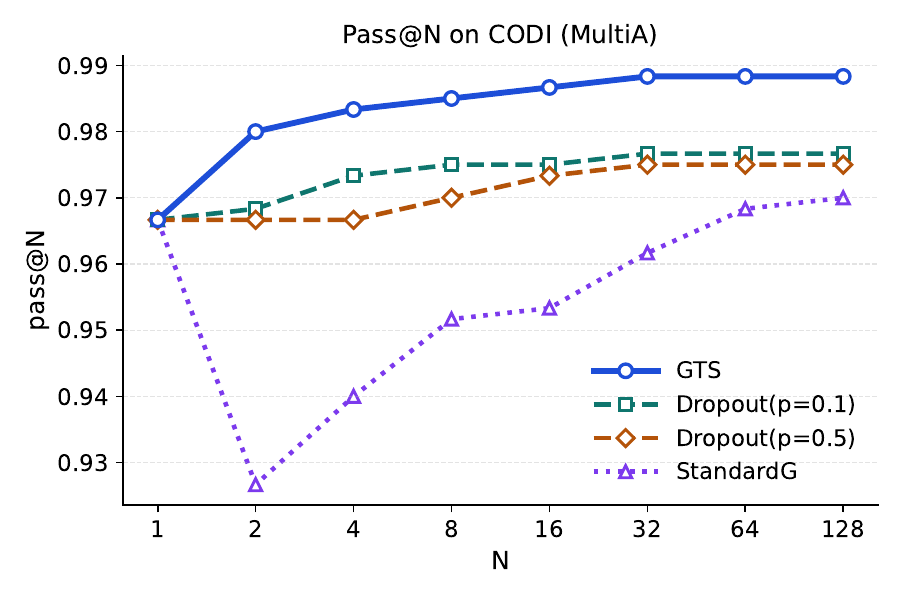} \\

            \includegraphics[width=0.48\textwidth]{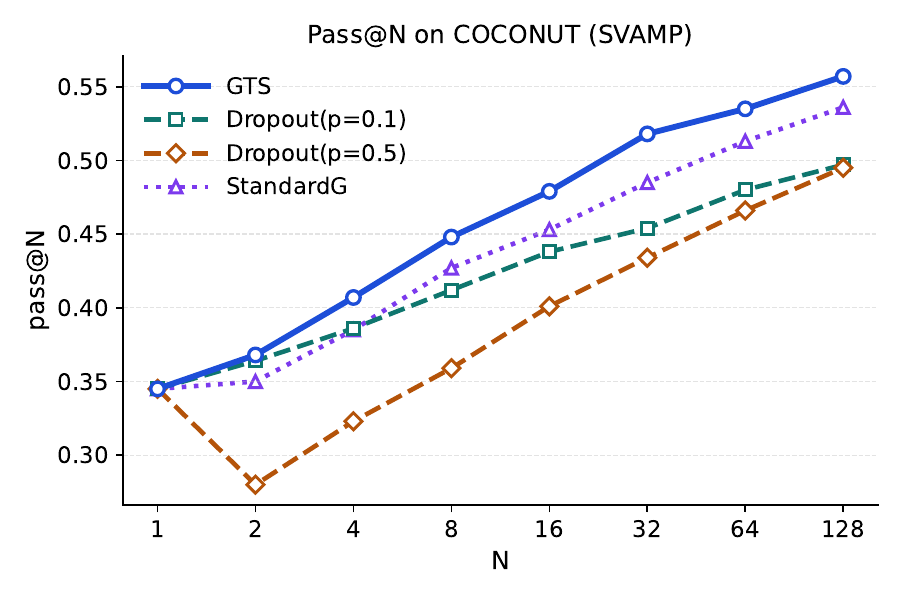} &
            \includegraphics[width=0.48\textwidth]{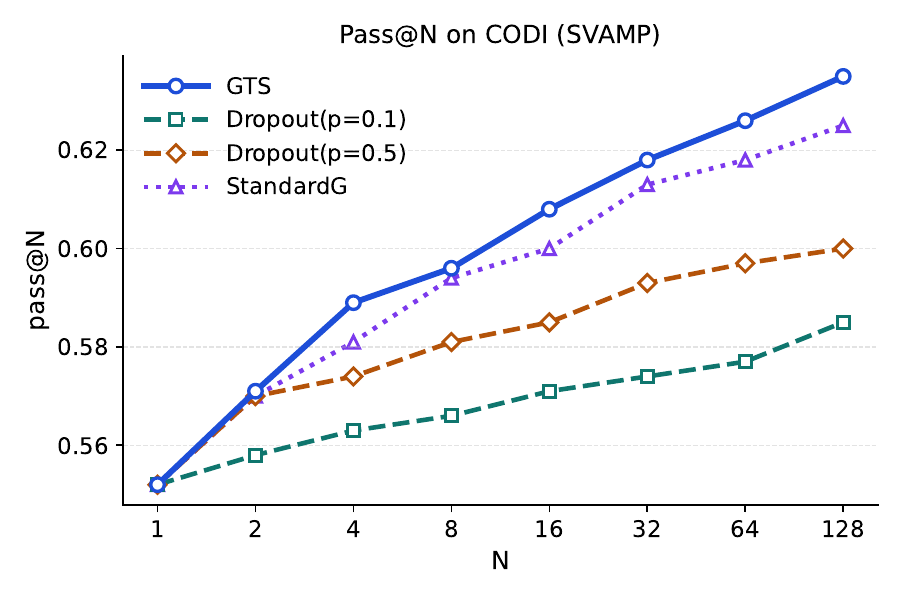} \\

            \includegraphics[width=0.48\textwidth]{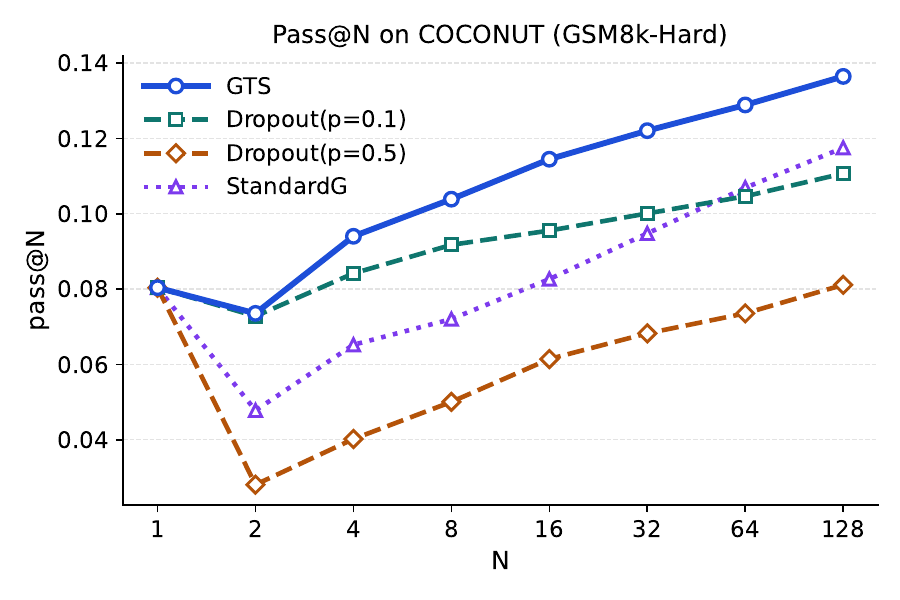} &
            \includegraphics[width=0.48\textwidth]{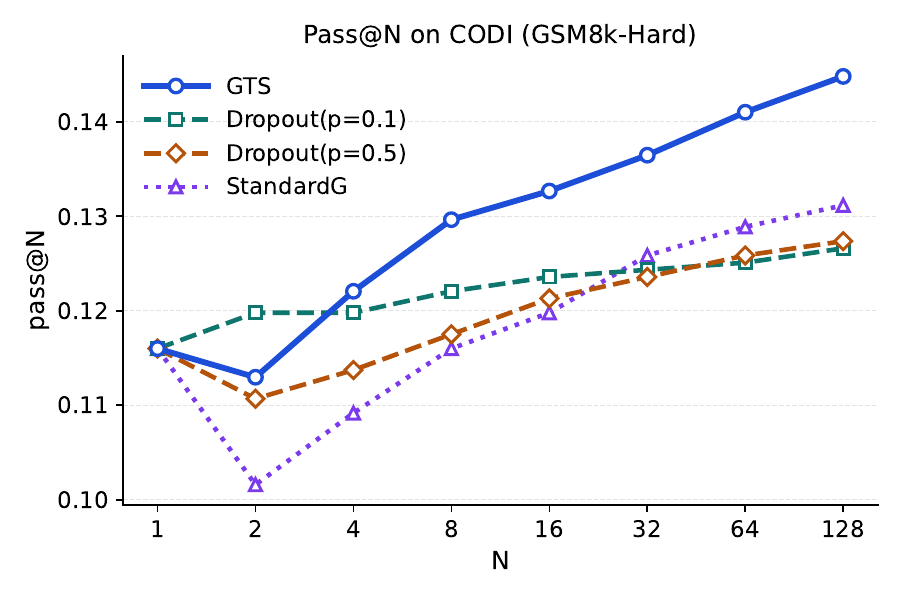}
        \end{tabular}
    }
    \caption{
     Out-of-distribution ITS performance on three benchmarks. Rows correspond to MultiArith, SVAMP, and GSM8K-Hard respectively, and columns correspond to \textsc{COCONUT} (left) and \textsc{CODI} (right).
    }
    \label{fig:latent_pan_ood}
\end{figure*}
\subsection{Out of Distribution Evaluation}
\label{app:ood_evaluation}

To examine whether the learned exploration policy generalizes beyond the training distribution, we evaluate the samplers trained on GSM8K-Aug on three out-of-distribution arithmetic reasoning benchmarks: MultiArith~\citep{roy-roth-2015-solving} (600 samples), SVAMP~\citep{patel-etal-2021-nlp} (1000 samples), and GSM8K-Hard~\citep{gao2023palprogramaidedlanguagemodels} (1319 samples). No additional training or hyperparameter tuning is performed for these datasets; the samplers are applied directly at inference time under the same settings used in the main experiments in~\Cref{sec:exp:setup}.

\Cref{fig:latent_pan_ood} reports the resulting pass@N curves for both \textsc{COCONUT} and \textsc{CODI}. Across all three benchmarks, GTS consistently achieves stronger scaling behavior than heuristic baselines, including dropout-based sampling and standard Gaussian perturbations. The improvement is particularly clear at moderate-to-large sampling budgets, indicating that the learned perturbation policy transfers more consistently to unseen arithmetic benchmarks and continues to yield stronger scaling behavior than heuristic baselines.

In contrast, heuristic sampling methods exhibit noticeably inconsistent behavior across benchmarks. The same perturbation configuration can perform reasonably well on one dataset but degrade significantly on another (e.g., CODI-StandardG on MultiA and SVAMP). This variability suggests that fixed heuristic perturbations are sensitive to differences in problem difficulty, input distribution, and task characteristics. Overall, these results further support the central motivation of this work: effective latent inference-time scaling requires better-controlled, context-aware sampling rather than globally fixed stochastic perturbations.

\section{Additional Details about GTS}
\subsection{Reward Shaping}
\label{app:reward_shaping}

For each input $x$, we sample a group of $N$ latent perturbation trajectories and obtain $N$ decoded answers.
Let $a^{(i)}$ denote the $i$-th decoded answer, and let
$\mathbb{I}[a^{(i)}=y^\star]$ be the exact-match indicator with respect to the ground-truth answer $y^\star$.
The reward for trajectory $i$ is defined as
\begin{equation}
r^{(i)}
\;=\;
r_0\,(2\,\mathbb{I}[a^{(i)}=y^\star]-1)
\;+\;
\alpha\, s^{(i)},
\label{eq:reward_appendix}
\end{equation}
where $r_0>0$ controls the base correctness magnitude and $\alpha$ scales a lightweight shaping term.

\paragraph{Base Correctness Term}
The first term assigns $+r_0$ to correct answers and $-r_0$ to incorrect ones.
This symmetric formulation ensures that correctness remains the dominant optimization signal.
When $\alpha=0$, the objective reduces to accuracy-only reward.

\paragraph{Confidence Score}
For each trajectory, we compute a scalar confidence score $c^{(i)}$ using the length-normalized log-probability of the generated answer:
\begin{equation}
c^{(i)} \;=\;
\frac{1}{|a^{(i)}|}
\sum_{t=1}^{|a^{(i)}|}
\log p_\theta\!\big(a^{(i)}_t \mid x, \tau^{(i)}, a^{(i)}_{<t}\big),
\end{equation}
where $|a^{(i)}|$ denotes answer length and $\tau^{(i)}$ is the sampled latent trajectory.
Length normalization prevents longer answers from being systematically penalized.

\paragraph{Within-Group Normalized Shaping}
To avoid directly optimizing raw likelihood (which could dominate correctness),
we apply group-wise normalization separately over correct and incorrect subsets.
Define
\begin{align}
    &\mathcal{C}=\{i:\mathbb{I}[a^{(i)}=y^\star]=1\}, \\
    \quad
    &\mathcal{W}=\{i:\mathbb{I}[a^{(i)}=y^\star]=0\}.
\end{align}
Within each subset (when its size is at least 3),
we compute a $z$-score normalization:
\begin{equation}
    z(c^{(i)}) \;=\;
    \frac{c^{(i)}-\mu_{\mathcal{S}}}{\sigma_{\mathcal{S}}},
    \quad
    \mathcal{S}\in\{\mathcal{C},\mathcal{W}\},    
\end{equation}

where $\mu_{\mathcal{S}}$ and $\sigma_{\mathcal{S}}$ are the mean and standard deviation of $c^{(i)}$ within the subset.
The shaping term is then defined as
\begin{equation}
s^{(i)}=
\begin{cases}
\ \tanh\!\big(\frac{z(c^{(i)})}{\tau}\big)
& i\in\mathcal{C},\ |\mathcal{C}|\ge 3 \\[6pt]
-\tanh\!\big(\frac{z(c^{(i)})}{\tau}\big)
& i\in\mathcal{W},\ |\mathcal{W}|\ge 3 \\[6pt]
0
& \text{otherwise}.
\end{cases}
\label{eq:shape_appendix}
\end{equation}
The temperature parameter $\tau$ controls the smoothness of the shaping signal, which is set as 1.0 in our experiments.

\paragraph{Design Rationale}
This construction has three desirable properties:
(1) correctness remains the primary optimization objective via $r_0$;
(2) shaping only provides relative discrimination \emph{within} correct and incorrect groups, rather than encouraging unconditional likelihood maximization; and (3) the $\tanh$ squashing bounds the shaping magnitude, ensuring it remains secondary to the correctness term.
Empirically, this dense signal improves within-group ranking of sampled trajectories without destabilizing training.

\subsection{Implementation Details}
\label{app:implementation_details}

\paragraph{Sampler Architecture}
GTS consists of two lightweight heads that predict the mean $\mu_\phi(c_k)$ and log standard deviation $\log \sigma_\phi(c_k)$ of a diagonal Gaussian policy at each latent step.
Each head is implemented as a two-layer feed-forward network with hidden dimension $D$, SiLU activation, and output dimension $D$.
Only the sampler parameters are trainable; all backbone parameters remain frozen.

To prevent premature collapse toward a deterministic policy, we clamp the minimum log standard deviation to $-2.0$, corresponding to $\sigma_{\min}=\exp(-2)\approx0.135$.
This lower bound ensures a non-trivial exploration scale throughout training.

For COCONUT, perturbations are directly added to the latent hidden state.
For CODI, which contains an additional recurrent filtering module with layer normalization at the end, perturbations are injected \emph{after} the recurrent filter output.
This placement ensures that stochastic perturbations are not attenuated by normalization layers and can effectively propagate to subsequent reasoning steps.

In terms of parameter size, GTS introduces approximately 2.3M parameters for COCONUT (1.8\% of the backbone) and 16M parameters for CODI (1.2\% of the backbone).

\paragraph{Perturbation Schedule Across Latent Steps}
Both models employ $K=6$ latent reasoning steps.
However, perturbations are injected only from \texttt{<start\_latent>} through \texttt{<lat\_5>}.

The final latent token \texttt{<lat\_6>} influences the output only through attention to \texttt{<end\_latent>} and does not re-enter the latent autoregressive loop.
Specifically, the subsequent \texttt{<end\_latent>} token is provided via teacher forcing when predicting the answer prefix, meaning that \texttt{<lat\_6>} affects decoding through key-value cache interactions but is not recursively fed back as a new latent state.
Injecting perturbations at \texttt{<lat\_6>} would therefore not fully propagate through the reasoning dynamics.
To ensure that stochasticity consistently influences autoregressive latent refinement, we restrict perturbations to the first five latent steps.

\paragraph{Policy Optimization Details}
The reference sampler $q_{\phi_{\text{ref}}}$ is updated as an exponential moving average (EMA) of the current policy with decay rate $0.999$.
The KL regularization coefficient is set to $\beta=0.001$.

Advantages are normalized at the prompt level: for each prompt, the $N=32$ rollout rewards are standardized before computing the policy objective.
With batch size 32 prompts, each optimization step processes $32 \times 32$ sampled trajectories jointly.

When computing trajectory log densities, a diagonal Gaussian formally requires summation over dimensions.
However, summing over high-dimensional latent vectors can produce large-magnitude log-density values, leading to unstable density ratios.
To improve numerical stability, we average over dimensions instead of summing.
This modification preserves relative likelihood ordering while keeping ratio magnitudes well-scaled for optimization.

For the clipped GRPO objective, the density ratio is clipped to the range $[-20, 20]$.
Because trajectory log-probabilities are accumulated across latent steps (rather than computed at a single step level), their scale differs from standard token-level PPO formulations.
The wider clipping interval empirically stabilizes training without restricting useful policy updates.

\paragraph{Training Configuration}
Unless otherwise stated, GTS is trained for 10K optimization steps with learning rate $1\times10^{-4}$ and linear warmup.
The reward shaping coefficient is $\alpha=0.2$.
All experiments are conducted on a single NVIDIA A100 GPU.

\subsection{Design Choices for the Sampling Policy}

\paragraph{Diagonal Gaussian Policy}
We adopt a diagonal Gaussian parameterization for the perturbation distribution.
This choice provides a favorable trade-off between expressiveness and stability.
A diagonal policy allows dimension-wise scaling and directional steering while preserving closed-form log-density and KL divergence expressions, which are essential for stable GRPO optimization.
In contrast, a full-covariance Gaussian would introduce $\mathcal{O}(D^2)$ parameters and substantially increase both memory cost and numerical instability, especially in high-dimensional latent spaces.
Given that the backbone representations already encode rich cross-dimensional correlations, a diagonal perturbation distribution is sufficient to provide flexible yet controllable exploration.

\paragraph{Additive Perturbation Formulation}
We model latent exploration through additive perturbations $\tilde{h}_k = h^{\text{det}}_k + z_k$.
This formulation preserves the original backbone dynamics and ensures that perturbations act as local steering signals rather than replacing latent representations.
Additive noise also yields a unit Jacobian transformation, allowing the perturbation density to be directly interpreted as a distribution over latent thought states.
More complex transformations (e.g., multiplicative gating or learned flows) could increase flexibility but would entangle exploration with backbone dynamics and complicate policy density computation.

\paragraph{Relation to Dropout-Based Bayesian Sampling}

Dropout has been interpreted as approximate Bayesian inference, where Bernoulli masking corresponds to a variational approximation over model weights and enables predictive uncertainty estimation via Monte Carlo sampling~\citep{gal2016dropoutbayesianapproximationrepresenting}.
Under this perspective, stochastic forward passes primarily serve to quantify epistemic uncertainty and improve calibration in weight space.
ITS in latent reasoning, however, constitutes a search problem: the objective is to increase the probability of obtaining at least one correct reasoning trajectory under a fixed sampling budget.
Accurate posterior uncertainty estimation does not necessarily imply optimal trajectory-level exploration for decision correction.
Our method therefore does not aim to approximate a weight posterior, but instead learns a reward-aligned perturbation policy over latent states tailored to the inference-time objective.

\section{Additional Analysis}

\subsection{Further Discussion on Main Results}
\label{app:further_main_results}

Beyond absolute pass@N improvements, we observe that the \emph{relative gains} brought by GTS differ across backbone architectures.
On \textsc{COCONUT}, the improvement spans a wider margin (approximately $35\!\rightarrow\!55$), whereas on \textsc{CODI} the gains are more moderate (approximately $48\!\rightarrow\!58$).
We hypothesize that this discrepancy reflects structural differences in model scale and latent representation geometry.

\paragraph{Model Scale and Latent Dimensionality}
\textsc{COCONUT}, built on a GPT-2 backbone, operates in a lower-dimensional latent space ($D=768$) with fewer layers.
In such a regime, perturbations added to latent states can propagate more directly through subsequent reasoning steps, allowing moderate steering signals to produce visible changes in downstream predictions.
By contrast, \textsc{CODI} employs a deeper architecture with substantially higher dimensional latent representations ($D=2048$).
In higher-dimensional spaces, meaningful directional steering becomes inherently more challenging: the action space grows with $D$, while the relative magnitude of any single perturbation component diminishes.
Moreover, deeper networks possess stronger internal correction dynamics, which can dampen or redistribute injected perturbations across layers.
As a result, while GTS remains effective on \textsc{CODI}, its relative gains are naturally smaller than those observed on the lower-dimensional backbone.

\paragraph{Sampler Capacity Relative to Backbone Size}
Although the sampler architecture adopts the same two-layer design in both settings, its \emph{relative capacity} differs with respect to the backbone.
As reported in~\Cref{app:implementation_details}, GTS introduces approximately 1.8\% additional parameters for \textsc{COCONUT} and 1.2\% for \textsc{CODI}.
While the absolute sampler size increases with latent dimension, the backbone grows more substantially in the higher-dimensional model.
Consequently, the sampler-to-backbone capacity ratio becomes smaller in \textsc{CODI}.

Moreover, the effective control problem scales with latent dimensionality.
A diagonal Gaussian policy in $D=2048$ dimensions operates over a substantially larger action space than in $D=768$, increasing the difficulty of learning precise context-conditioned steering directions.
The same architectural design therefore faces a more demanding control landscape in larger latent manifolds.

From this perspective, the reduced relative gain on \textsc{CODI} does not indicate diminished effectiveness of structured perturbation, but rather reflects the increased complexity of steering higher-dimensional reasoning dynamics.
Future work may explore model-specific sampler architectures, including deeper sampler heads or layer-wise perturbation mechanisms, to better match backbone scale.
In the present work, however, we deliberately keep the sampler lightweight to isolate and validate the core idea of learnable, controlled latent exploration.

\paragraph{On the Small-$N$ Behavior}
We also observe a mild performance drop at very small sampling budgets (e.g., $N=2$).
This phenomenon is not unique to GTS and reflects a general exploration-exploitation trade-off:
when only a few samples are available, even structured perturbations may temporarily disrupt otherwise correct deterministic reasoning.
As $N$ increases, the probability that at least one trajectory meaningfully improves the internal decision state grows rapidly, leading to the observed recovery and scaling gains.

Importantly, this small-$N$ degradation is not fundamental.
One could introduce an explicit budget-aware scaling factor on the perturbation magnitude (e.g., modulating $\mathbf{z}$ as a function of $N$) to suppress exploration at very small budgets.
We deliberately avoid such adaptive scheduling to ensure a fair comparison across methods and sampling budgets.
Designing budget-sensitive exploration control remains an interesting direction for improving ITS in continuous reasoning space.

\subsection{Further Discussion on Sampling Behavior}
\label{app:snr_discussion}

To better understand the structural differences observed in the step-wise SNR distributions (\Cref{fig:snr}), we relate them to the underlying training objectives and latent reasoning formulations of the two models.

\paragraph{\textsc{COCONUT}: Paired Latent-Step Dynamics}
For \textsc{COCONUT}, SNR exhibits a clear saw-tooth pattern across adjacent latent steps.
Median values remain in a moderate range (approximately $0.3$--$0.5$), indicating that deterministic steering and stochastic variation are comparable in magnitude throughout reasoning.

This alternating structure is consistent with \textsc{COCONUT}'s training design.
\textsc{COCONUT} is trained via curriculum learning that progressively compresses textual reasoning into latent thought representations.
Importantly, its original training formulation contains three effective reasoning steps, each composed of \emph{two} latent thought vectors.
Within each pair, the first sub-step primarily consolidates information, while the second refines or expands upon it.

The SNR plot suggests that GTS adapts to this internal structure.
The first sub-step within each effective reasoning pair tends to exhibit relatively stronger deterministic steering, while the second allows comparatively greater stochastic exploration.
Rather than imposing a rigid perturbation schedule, GTS could align with the base model's intrinsic reasoning rhythm, preserving the alternation between consolidation and variation.

\paragraph{\textsc{CODI}: Front-Loaded Deterministic Adjustment}
In contrast, \textsc{CODI} displays a front-loaded SNR profile.
Deterministic strength peaks at the second latent step and then transitions into a sustained lower-SNR regime for subsequent steps.
Later reasoning steps therefore operate in a more exploration-dominated setting.

This behavior is consistent with \textsc{CODI}'s training objective.
Unlike \textsc{COCONUT}, which compresses intermediate textual reasoning through staged curriculum learning, \textsc{CODI} relies on distillation to align latent reasoning with the text reasoning answer prefix at the final step.
As a result, its latent trajectory is trained to evolve more smoothly and coherently across steps, without an explicit paired-step structure.

Under this formulation, a single early deterministic adjustment may be sufficient to steer the trajectory toward a promising region of latent space, after which controlled exploration can proceed without further strong intervention.
The observed SNR profile therefore reflects how GTS adapts to the more continuous and globally aligned reasoning dynamics of \textsc{CODI}.

\paragraph{Adaptive Rather Than Prescriptive Control}
Taken together, these patterns indicate that GTS does not enforce a uniform perturbation schedule across architectures.
Instead, the balance between deterministic steering and stochastic variation emerges from interaction with each model's latent reasoning dynamics.
The SNR distributions thus suggest that context-dependent perturbations can adapt to model-specific latent dynamics, rather than uniformly amplifying or suppressing stochasticity across all reasoning steps.

\end{document}